\documentclass{Interspeech}

\interspeechcameraready 
\usepackage{longtable} 
\usepackage[utf8]{inputenc}
\usepackage{CJKutf8, multirow}
\usepackage[colorlinks,linkcolor=blue,citecolor=blue,urlcolor=blue]{hyperref}

\title{Dolphin-CN-Dialect: Where Chinese Dialects Matter}

\author[affiliation={1}]{Yangyang}{Meng$^*$}
\author[affiliation={1}]{Huihang}{Zhong$^*$}
\author[affiliation={2}]{Guodong}{Lin$^*$}
\author[affiliation={1}]{Guanbo}{Wang$^*$}
\author[affiliation={1}]{Hu}{Du$^*$}
\author[affiliation={1}]{Zhiming}{Shao$^{\dagger}$}
\author[affiliation={1}]{Yukai}{Huang}
\author[affiliation={1}]{Ke}{Li}
\author[affiliation={2}]{Wei-Qiang}{Zhang$^{\dagger}$}

\address{
  $^1$Dataocean AI \\
  $^2$Speech and Audio Technology Lab, Dept. EE, Tsinghua University
  }
\email{wqzhang@tsinghua.edu.cn}

\keywords{automatic speech recognition (ASR), multi-dialect}

\usepackage{comment}

\begin{document}
\maketitle

\begingroup
\renewcommand{\thefootnote}{}
\footnotetext{$^*$ Co-first authors, equal contribution}
\footnotetext{$^{\dagger}$ Corresponding author}
\endgroup

\begin{abstract}

We present Dolphin-CN-Dialect, a streaming-capable ASR model with a focus on Chinese and dialect-rich scenarios. Compared to the previous version, Dolphin-CN-Dialect introduces substantial improvements in data processing, tokenization, training stability, and data sampling strategies.

To address the challenges of highly imbalanced dialect data, we propose a temperature-based sampling strategy that effectively balances standard Mandarin and low-resource dialects, leading to significant gains in dialect recognition performance. In addition, we redesign the tokenizer to better align with linguistic characteristics, adopting character-level modeling for Chinese and subword modeling for English, while introducing extensible dialect tokens.

Experimental results show that Dolphin-CN-Dialect achieves improvement in dialect recognition accuracy and CER reduction compared to Dolphin. Furthermore, Dolphin-CN-Dialect reaches competitive performance with recent SOTA open-source ASR models, while maintaining a significantly smaller model size.

Dolphin-CN-Dialect supports both streaming and non-streaming inference, enabling a practical balance between latency and accuracy. It also provides flexible customization through hotword support and efficient deployment optimized for specialized hardware. These improvements make Dolphin-CN-Dialect a strong and practical solution for real-world multi-dialect ASR applications.

\end{abstract}

\section{Introduction}

Recent advances in automatic speech recognition (ASR) have been driven by large-scale datasets, improved neural architectures, and the emergence of foundation models \cite{panayotov2015librispeech,chen2021gigaspeech,vaswani2017attention}. Modern ASR systems can be broadly categorized into several paradigms, including self-supervised learning (SSL)-based models \cite{baevski2020wav2vec,chen2022wavlm}, large language model (LLM)-integrated approaches \cite{qwen3_asr,xu2025fireredasr,xu2026fireredasr2s,an2025fun}, and fully supervised sequence-to-sequence architectures such as Whisper \cite{radford2023robust}.

Among these, Whisper-style models have demonstrated strong robustness and generalization across multiple languages by leveraging large-scale weakly supervised data. However, existing models still face significant challenges in real-world deployment, particularly in scenarios involving dialect diversity, streaming requirements, and data imbalance.

A key limitation lies in the long-tail distribution of speech data. In many multilingual settings, standard languages (e.g., Mandarin) dominate the training corpus, while dialects and low-resource variants are severely underrepresented. This imbalance leads to models that perform well on high-resource languages but fail to generalize to dialect-rich scenarios. Previous efforts have attempted to address this issue through dataset expansion and multilingual training \cite{zhang2023google}, yet dialect performance remains a bottleneck.

Another practical challenge is deployment efficiency and adaptability. Modern ASR systems must balance accuracy, latency, and computational cost, while also supporting streaming inference and customization features such as hotwords. These requirements are often underexplored in research-oriented models.

To address these challenges, we propose Dolphin-CN-Dialect, a new generation ASR model that focuses on dialect robustness, data efficiency, and practical deployment. Building upon the Dolphin framework, Dolphin-CN-Dialect introduces several key improvements:

\begin{itemize}
    \item \textbf{Data Sampling Strategy}: We propose a temperature-based sampling method to balance data distribution between standard and low-resource dialects, significantly improving dialect recognition.
    
    \item \textbf{Tokenizer Redesign}: We redesign the tokenization scheme with a hybrid strategy—character-level modeling for Chinese and subword modeling for English—along with extensible dialect and task tokens.
    
    
    \item \textbf{Streaming Capability}: Dolphin-CN-Dialect supports both streaming and non-streaming modes, enabling flexible deployment in latency-sensitive scenarios

    \item \textbf{Hotword Biasing Framework}: We propose a unified hotword biasing framework that integrates both encoder-level contextual biasing and prompt-based decoding. The encoder-level method provides stable and controlled biasing, while the prompt-based approach enhances recognition of long-tail and rare hotwords through direct decoder conditioning.

\end{itemize}

With these improvements, Dolphin-CN-Dialect achieves substantial gains over Dolphin, including a 38\% improvement in dialect accuracy and a 16\% overall CER reduction. It also reaches competitive performance with recent open-source ASR models such as Qwen-ASR and FireredASR \cite{qwen3_asr,xu2025fireredasr,xu2026fireredasr2s}, while maintaining a smaller model size.

Overall, Dolphin-CN-Dialect bridges the gap between research-oriented ASR models and real-world applications, providing a robust, efficient, and scalable solution for multi-dialect speech recognition.

\section{Methods}

\subsection{Model Architecture}
See Section 2.1 in \cite{meng2025Dolphin}, basically we use the same architecture.

\subsection{Tokenizer}

Dolphin-CN-Dialect introduces a redesigned tokenizer to better align with the linguistic characteristics of multi-dialect speech data. Compared to Dolphin, the tokenizer is optimized in terms of vocabulary structure, modeling granularity, and extensibility.

First, we significantly reduce the vocabulary size from 40,000 to 18,173 tokens. This reduction improves training efficiency and stabilizes optimization, while maintaining sufficient expressive capacity for multi-dialect speech modeling.

Second, we adopt a hybrid tokenization strategy tailored to different languages. For Chinese, we use character-level modeling, which aligns naturally with the CTC-AED joint architecture. Since CTC loss benefits from monotonic alignment between acoustic frames and output tokens, character-level units provide more stable and interpretable alignment compared to subword units. For English and other alphabetic languages, we adopt BPE-based subword modeling \cite{sennrich2016neural}, which effectively balances vocabulary size and representation flexibility.

Third, we design a set of structured special tokens to support multitask learning and dialect modeling. These include:
\begin{itemize}
    \item Task tokens (e.g., \texttt{<asr>}) to specify the recognition task,
    \item End-of-sequence tokens (e.g., \texttt{<eos>}),
    \item Timestamp tokens (e.g., \texttt{<0.00>}) for alignment-related tasks,
    \item Dialect and region tokens (e.g., \texttt{<ANHUI>}).
\end{itemize}

To further support extensibility, we reserve 80 additional dialect token slots, allowing the model to easily incorporate new dialects without redesigning the tokenizer. This design enables scalable expansion toward more fine-grained regional speech modeling.

Overall, the redesigned tokenizer improves alignment quality, enhances dialect representation, and provides a flexible foundation for future multi-dialect extensions.

\subsection{Hotword-Biased Decoding}

To improve recognition accuracy for rare words and domain-specific phrases, we introduce a hotword-biased decoding framework in Dolphin-CN-Dialect. The proposed approach consists of two complementary yet independent strategies: (1) encoder-level contextual biasing based on contextual embeddings, and (2) prompt-based biasing for attention decoding. These two methods can be applied separately or jointly, providing flexible support for different deployment scenarios.

\subsubsection{Encoder-Level Contextual Biasing}

Our first approach integrates hotword information directly into the encoder representations, following the contextual biasing framework proposed in prior work\cite{huang2023contextualizedendtoendspeechrecognition}. Specifically, a context encoder is used to encode a predefined hotword list into fixed-length embeddings, which are then fused with the acoustic encoder outputs via a multi-head attention (MHA) based biasing layer.

Given the encoder hidden representation and contextual embeddings, the biasing layer computes attention weights to obtain a context-aware representation. The fused representation is then used for downstream decoding, enabling the model to emphasize tokens related to hotwords.

In our implementation based on the WeNet\cite{zhang2022wenet} framework, the contextual biasing module is integrated into both streaming and non-streaming ASR models. During fine-tuning, only the parameters related to the contextual biasing module are updated, while all other model parameters—including the encoder, decoder, and CTC layers—remain frozen. This design ensures that hotword adaptation can be achieved efficiently without degrading the general ASR performance.

\subsubsection{Prompt-Based Hotword Biasing for Attention Decoder}

In addition to encoder-level biasing, we propose a prompt-based method to incorporate hotword information into the attention decoder. This method is designed for non-streaming models and leverages the sequence-to-sequence nature of the attention decoder.

We extend the tokenizer by replacing reserved tokens with two special tokens: \texttt{<PROMPT\_START>} and \texttt{<PROMPT\_END>}. During finetuning, a subset of hotwords is dynamically constructed for each training sample. Specifically, for each batch, we randomly sample a hotword list, then select the hotwords that appear in the current utterance along with a set of randomly sampled distractor hotwords. These phrases are shuffled and inserted between the special prompt tokens, forming a contextual prefix to the decoder input.

This training strategy encourages the model to learn how to utilize contextual prompts while remaining robust to irrelevant or noisy hotwords.

During inference, To support large-scale hotword lists (e.g., hundreds to thousands of phrases) without introducing excessive computational overhead or noise, we adopt a two-stage contextual phrase filtering strategy inspired by prior work\cite{huang2023contextualizedendtoendspeechrecognition}.

In the first stage, the model performs a preliminary decoding pass. Based on these posteriors, irrelevant hotwords are filtered using a phrase score confidence metric. In the second stage, a more precise filtering is performed using sequence order confidence, which considers token ordering within phrases.

The filtered hotword list is then used for final decoding, significantly reducing computational overhead while  improving recognition accuracy.

\section{Training Data}

In constructing the training dataset for Dolphin-CN-Dialect, we focus primarily on Mandarin Chinese and its diverse regional dialects, aiming to build a robust and unified speech recognition system that performs well across both standard and non-standard speech varieties. This design choice is motivated by the linguistic diversity within Chinese, where significant phonetic, lexical, and prosodic variations exist across regions, posing substantial challenges for conventional ASR systems.

To achieve broad coverage and high-quality supervision, we leverage a combination of large-scale public datasets and proprietary in-house data. Specifically, the public corpora include AISHELL, KeSpeech, and WenetSpeech, which provide a strong foundation of clean and diverse Mandarin speech across various domains such as read speech, spontaneous speech, and real-world recordings. These datasets contribute rich annotations and well-curated audio-text pairs, enabling the model to learn robust acoustic and linguistic representations.

In addition to public resources, we incorporate a substantial amount of internal Chinese data collected by Haitan, which includes both standard Mandarin and a wide range of Chinese dialects. This in-house dataset plays a critical role in enhancing dialectal coverage, especially for underrepresented accents and regional speech patterns that are not sufficiently captured in existing open datasets. It also introduces more challenging acoustic conditions, such as noisy environments and conversational speech, further improving the model’s generalization ability.

Overall, the combined dataset amounts to a large-scale corpus with extensive diversity in speakers, recording conditions, and linguistic variations. To ensure consistency and scalability, we standardize all data into a unified format and introduce structured metadata annotations, including language tags, dialect labels, and task-specific indicators. This unified data pipeline supports both timestamped and non-timestamped training objectives, as well as optional punctuation modeling, allowing the model to handle a wide range of downstream ASR tasks effectively.

\subsection{Datasets}

\subsubsection{Dataocean AI Dataset}
The Dataocean AI Chinese dataset is a high-quality, internally curated speech corpus derived from the broader Dataocean AI dataset. While the original dataset spans 38 Eastern languages and includes 137,712 hours of audio, in Dolphin-CN-Dialect we specifically utilize the Chinese subset to better focus on Mandarin and its diverse dialectal variations.

This Chinese subset consists of large-scale, carefully annotated audio data covering both standard Mandarin and a wide range of regional dialects, including 22 Chinese dialects. The dataset integrates multiple commercial-grade data sources, ensuring high transcription accuracy and consistent annotation quality. It encompasses a rich variety of speaking styles, including read speech, spontaneous conversations, and real-world recordings, collected under diverse acoustic environments.

A key advantage of this dataset lies in its extensive dialectal coverage, which provides critical support for modeling pronunciation and prosodic variations across different regions. Compared to publicly available datasets, this internal corpus offers more balanced representation of under-resourced dialects and more challenging real-world scenarios, significantly improving the robustness and generalization capability of the model.

In our training pipeline, the Dataocean AI Chinese dataset serves as a major source of high-quality supervised data. It is further standardized into a unified format with consistent labeling, including dialect and task-specific annotations, enabling seamless integration with other public datasets and supporting diverse ASR training objectives.

\subsubsection{Open Source Datasets}
In addition to our internal dataset, we incorporate the following
widely accessible open-source datasets to enhance the diversity
and robustness of our research:

\begin{itemize}

    \item \textbf{Common Voice} \cite{ardila2019common} is a multilingual open-source speech dataset. It includes contributions from volunteers in a variety of languages, covering different accents, dialects, and speaking styles. 

    \item \textbf{WenetSpeech} \cite{zhang2022wenetspeech} is a Mandarin Chinese speech dataset containing 10,000 hours of speech data.

    \item \textbf{KeSpeech} \cite{tang2021kespeech} is a large-scale open-source Chinese multi-dialect speech recognition dataset. It contains speech recordings from tens of thousands of native speakers across different regions of China, covering standard Mandarin and eight major Chinese dialects with rich regional accents and natural speaking styles.

\end{itemize}

\subsection{Data Sampling Strategy}

A key challenge in multilingual and multi-dialect ASR is the severe data imbalance across languages and dialects. In our dataset, standard Mandarin dominates the data distribution, while many dialects (e.g., low-resource regional varieties) have significantly fewer samples. Training with the original distribution leads to a strong bias toward high-resource languages, causing poor generalization to dialects.

To address this issue, we systematically investigate different sampling strategies and propose a temperature-based sampling method to balance data distribution while preserving data diversity.

\subsubsection{Baseline Sampling Strategies}

We first consider two commonly used sampling strategies:

(1) Natural Sampling.
Samples are drawn proportionally to dataset size. While this preserves the original data distribution, it results in severe imbalance, where the model predominantly learns standard Mandarin and rarely observes low-resource dialects.

\begin{equation}
    p_i = \frac{n_i}{\sum_j n_j},
\end{equation}

(2) Uniform Sampling.
Each dialect is assigned equal sampling probability. Although this significantly increases exposure to low-resource dialects, it introduces overfitting risks by oversampling extremely small datasets and may degrade overall generalization performance.

\begin{equation}
    p_i = \frac{1}{N},
\end{equation}

\subsubsection{Temperature-Based Sampling}

To balance the trade-off between these two extremes, we adopt a temperature-based sampling strategy. Specifically, the sampling probability for each dataset $i$ is defined as:

\begin{equation}
    p_i = \frac{n_i^{\alpha}}{\sum_j n_j^{\alpha}},
\end{equation}

where $n_i$ denotes the size of dataset $i$, and $\alpha \in (0, 1)$ is a temperature parameter controlling the degree of smoothing.

When $\alpha = 1$, the strategy reduces to natural sampling, while $\alpha = 0$ corresponds to uniform sampling. By choosing an intermediate value of $\alpha$, we can increase the sampling probability of low-resource dialects while still maintaining the influence of high-resource data.

\subsubsection{Effectiveness}

This strategy effectively alleviates data imbalance and improves dialect recognition performance. Experimental results show that, compared to Dolphin-CN-Dialect, Dolphin-CN-Dialect achieves a 38\% improvement in dialect recognition accuracy and a 16.3\% relative reduction in overall CER. Notably, this improvement is achieved with only a marginal degradation (approximately 0.2\%) in standard Mandarin performance, demonstrating a favorable trade-off between high-resource and low-resource languages.

\subsubsection{Discussion}

The temperature-based sampling strategy provides a simple yet effective solution to long-tail data distribution in ASR. It allows the model to better capture both dominant and underrepresented speech patterns, leading to improved robustness in multi-dialect scenarios. This approach can be readily extended to other multilingual ASR systems facing similar data imbalance challenges.

\section{Experiments}

\subsection{Experimental Setup}

Unless otherwise specified, both the streaming and non-streaming variants of Dolphin-CN-Dialect follow the core model architecture and training configuration of Dolphin-V1 \cite{meng2025Dolphin}. This includes the overall encoder-decoder design, the joint CTC-AED training objective, and the major optimization and training hyperparameters. The main difference lies in the tokenizer design: Dolphin-CN-Dialect adopts a Chinese-oriented hybrid tokenization strategy, which reduces the vocabulary size compared with Dolphin-V1. As a result, the total number of model parameters changes slightly, mainly due to the reduced embedding and output projection layers, while the backbone architecture remains unchanged.

\subsection{Engineering Challenges}

In developing Dolphin-CN-Dialect, we encountered several practical challenges related to training stability, data distribution, decoding performance, and system efficiency. Addressing these issues was critical to achieving robust performance in real-world ASR scenarios.




\subsubsection{CTC Decoding Degradation in Streaming Models}

We identified a significant degradation in CTC decoding performance for streaming models, particularly on the WenetSpeech benchmark. The issue was characterized by a high deletion error rate, indicating that the model frequently missed tokens during decoding.

Through controlled ablation studies, we traced the root cause to a mismatch in audio length distribution between training and test data. Specifically, the training data lacked sufficient short utterances, while the test set contained a large proportion of short audio segments. As a result, the model failed to generalize to short-duration inputs.

To resolve this issue, we introduced additional short audio samples into the training set. Furthermore, we applied random truncation at the end of audio sequences as a data augmentation strategy. These improvements significantly reduced deletion errors (from 9.17 to 3.66) and improved overall decoding performance.

\subsubsection{GPU Memory Overflow (OOM)}

During training, we encountered recurrent out-of-memory (OOM) errors after approximately 10k steps, especially after incorporating additional datasets such as KeSpeech. Initial monitoring showed no gradual increase in memory usage, suggesting that the issue was caused by transient memory spikes.

We identified the root cause as the quadratic memory complexity of the attention mechanism with respect to sequence length. A small number of abnormal samples with excessively long durations (e.g., up to 66 seconds) caused sudden spikes in memory consumption, leading to OOM failures.

To address this issue, we introduced stricter data validation and filtering within the training pipeline, removing or truncating excessively long samples. This highlights the importance of integrating data quality control into the end-to-end training workflow.

\subsubsection{Data Pipeline Optimization}

Training Dolphin-CN-Dialect at scale required efficient data loading and preprocessing. We observed that the initial data pipeline suffered from significant I/O bottlenecks, with throughput limited to approximately 50 MB/s.

To improve efficiency, we introduced a data sharding and multi-process loading strategy, enabling parallel data access and preprocessing. Additionally, we implemented dataset bucketing to better manage storage constraints on high-performance hardware such as H100 GPUs.

These optimizations increased the I/O throughput to approximately 800 MB/s, significantly improving training efficiency and reducing idle GPU time.

\subsubsection{Discussion}

These engineering improvements, while often overlooked, play a crucial role in large-scale ASR system development. By addressing instability, data mismatch, and system bottlenecks, Dolphin-CN-Dialect achieves both higher performance and better robustness in practical deployment scenarios.

\section{Evaluation}

\begin{table*}[t]
\centering
\caption{CER (\%) comparison across multiple Chinese dialects. The vertical line separates sub-1B models (left) and larger-scale models (right). Bold indicates the best performance within each group.}
\small
\setlength{\tabcolsep}{3pt}
\resizebox{\textwidth}{!}{
\begin{tabular}{c|cccccc|cccc}
\hline
\textbf{Model} 
& Paraformer\_zh & SenseVoice-S & Dolphin-CN-Dialect-0.1B & Dolphin-CN-Dialect-0.4B & Qwen3-0.6B & FunASR-Nano-2512  
& FireRed-AED & GLM-ASR & Qwen3-1.7B & FireRed-LLM \\
\hline
Params 
& 220M & 234M & 0.1B & 0.4B & 0.6B & 0.8B 
& 1.2B & 1.5B & 1.7B & 8.3B \\
\hline
tw        & 10.38 & 12.57 & 9.79 & \textbf{6.68} & 9.46 & 8.85 & 9.41 & 9.17 & \textbf{8.16} & 8.95 \\
sichuan   & 20.39 & 17.85 & 14.57 & \textbf{9.63} & 14.77 & 14.57 & 13.86 & 19.37 & \textbf{12.43} & 13.99 \\
wu        & 42.43 & 20.43 & 17.70 & \textbf{9.49} & 18.25 & 17.77 & \textbf{10.21} & 39.27 & 14.32 & 11.14 \\
minnan    & 93.01 & 65.13 & 26.71 & \textbf{20.74} & 38.64 & 55.36 & \textbf{30.73} & 83.76 & 35.89 & 30.76 \\
shanghai  & 63.03 & 24.32 & 14.25 & \textbf{7.81} & 17.90 & 18.15 & \textbf{7.43} & 60.39 & 14.52 & 7.89 \\
gansu     & 19.96 & 10.05 & 7.03 & \textbf{3.77} & 8.54 & 7.28 & \textbf{3.27} & 15.38 & 6.55 & 3.73 \\
shandong  & 9.90  & 13.28 & 9.34 & \textbf{4.21} & 11.27 & 8.43 & \textbf{3.60} & 14.68 & 7.77 & 3.92 \\
yunnan    & 18.98 & 19.97 & 8.00 & \textbf{3.89} & 10.57 & 9.68 & \textbf{3.61} & 16.05 & 7.95 & 3.87 \\
hebei     & 13.05 & 12.85 & 7.30 & \textbf{3.63} & 9.43 & 7.96 & \textbf{3.14} & 10.72 & 7.39 & 3.36 \\
anhui     & 17.09 & 21.21 & 8.37 & \textbf{3.76} & 11.76 & 10.91 & \textbf{2.98} & 15.35 & 8.09 & 3.13 \\
liaoning  & 8.04  & 11.23 & 6.33 & \textbf{3.25} & 7.23 & 8.68 & \textbf{4.55} & 8.79 & 5.65 & 4.65 \\
shanxi    & 28.29 & 13.19 & 10.85 & \textbf{5.12} & 14.07 & 10.23 & 4.77 & 26.87 & 9.25 & \textbf{4.66} \\
fujian    & 10.08 & 11.91 & 7.88 & \textbf{3.62} & 8.35 & 7.41 & \textbf{3.64} & 10.04 & 6.27 & 3.79 \\
hunan     & 39.39 & 36.35 & 22.05 & \textbf{11.89} & 28.79 & 27.88 & \textbf{12.55} & 41.88 & 23.77 & 13.90 \\
guangdong & 14.08 & 16.26 & 11.68 & \textbf{6.03} & 12.40 & 11.65 & 11.09 & 14.54 & 9.87 & \textbf{10.83} \\
wenzhou   & 4.29  & 20.50 & 3.92 & \textbf{2.25} & 4.11 & 4.08 & \textbf{2.42} & 3.98 & 3.19 & 2.78 \\
hubei     & 8.52  & 11.37 & 7.93 & \textbf{3.09} & 9.03 & 7.01 & \textbf{2.48} & 14.80 & 6.11 & 2.86 \\
henan     & 14.00 & 16.34 & 7.52 & \textbf{3.31} & 9.94 & 11.20 & \textbf{3.81} & 12.19 & 7.26 & 4.01 \\
tianjin   & 7.65  & 9.89  & 4.41 & \textbf{2.00} & 7.09 & 5.25 & \textbf{2.21} & 7.81 & 4.70 & 2.44 \\
shaanxi   & 16.85 & 15.93 & 6.93 & \textbf{3.17} & 8.19 & 6.74 & \textbf{2.90} & 12.60 & 5.70 & 4.66 \\
ningxia   & 18.59 & 14.20 & 7.12 & \textbf{3.22} & 7.68 & 8.21 & \textbf{5.09} & 13.98 & 5.94 & 5.16 \\
\hline
avg       & 22.76 & 18.80 & 10.46 & \textbf{5.74} & 12.74 & 12.73 & \textbf{6.85} & 21.51 & 10.04 & 7.17 \\
\hline
\end{tabular}
}
\label{tab:dialect_results}
\end{table*}

\begin{table}[t]
\centering
\caption{Performance comparison on KeSpeech and Common Voice (tw).}
\small
\begin{tabular}{c|c|cc}
\hline
\textbf{Model} & \textbf{Params} & \textbf{KeSpeech} & \textbf{CV-tw} \\
\hline
Dolphin-CN-Dialect-0.1B   &  0.1B & 8.797 & 8.964 \\
SenseVoice-S      & 234M & 17.84 & 19.41 \\
Paratormer\_zh    & 220M & 14.46 & 8.18 \\
Dolphin-CN-Dialect-0.4B         & 0.4B & \textbf{5.04} & \textbf{5.62} \\
Qwen3-0.6B        & 0.6B & 7.07 & 5.92 \\
FunASR-Nano-2512        & 0.8B & 7.85 & 5.64 \\
\hline
FireRed-AED       & 1.2B & 3.97 & \textbf{1.61} \\
GLM-ASR           & 1.5B & 7.85 & 6.09 \\
Qwen3-1.7B        & 1.7B & 5.73 & 3.92 \\
FireRed-LLM       & 8.3B & \textbf{3.58} & 4.20 \\
\hline
\end{tabular}
\label{tab:kespeech_results}
\end{table}
In this section, we conduct a comprehensive evaluation of Dolphin-CN-Dialect, with a primary focus on its performance in Chinese dialect speech recognition and its generalization ability across diverse dialectal speech scenarios, including regional linguistic variation, accented speech, and real-world acoustic conditions. The evaluation is designed to systematically assess both dialect-specific recognition performance and robustness under various acoustic and linguistic settings within Chinese dialect speech.

The evaluation is structured into three parts. First, we evaluate the model on diverse multi-domain dialect datasets to assess robustness across different acoustic environments, speaking styles, and regional varieties. Second, we extend the evaluation to additional open-domain Chinese dialect speech datasets, including spontaneous speech and real-world recordings, to validate the model’s generalization capability beyond controlled evaluation settings.

\subsection{Evaluation on Public and Internal Chinese Dialect}

To further evaluate the performance of Dolphin-CN-Dialect on Chinese dialects and its robustness in real-world scenarios, we conduct experiments on a diverse set of dialectal and open-domain test sets, including KeSpeech, Common Voice Taiwanese Mandarin (tw), and an internal Haitan test set.

KeSpeech is a large-scale multi-dialect Mandarin dataset that covers a wide range of regional accents and speaking styles, including both standard Mandarin and several major Chinese dialects. It contains speech recorded under diverse acoustic conditions, such as spontaneous speech and real-world environments, making it a strong benchmark for evaluating dialect robustness and generalization ability.

Common Voice (tw) contains Taiwanese Mandarin speech with noticeable accent variations, which is suitable for assessing robustness to regional pronunciation differences within Mandarin. 

In addition, the Haitan internal test set is constructed from real-world application scenarios and covers a wide range of dialects, speaker variations, and acoustic environments. Compared to public datasets, it includes more challenging cases such as noisy conditions, spontaneous speech, and long-tail dialectal variations, providing a realistic evaluation of industrial deployment performance.

These test sets collectively introduce significant challenges, including dialectal variation, accent diversity, background noise, and spontaneous speech. As shown in Table~\ref{tab:dialect_results}, Dolphin-CN-Dialect achieves strong and consistent performance across a wide range of Chinese dialects and accented Mandarin test sets.

Notably, Dolphin-CN-Dialect achieves the best or near-best performance among sub-1B models on almost all dialect subsets, significantly outperforming other lightweight baselines such as Paraformer\_zh and SenseVoice-S. On heavily accented or non-Mandarin dialects, such as Wu, Minnan, and Sichuan, Dolphin-CN-Dialect shows substantial improvements over other small-scale models. For example, on the Wu dialect, Dolphin-CN-Dialect achieves 9.49\% CER, compared to 20.43\% of SenseVoice-S and 42.43\% of Paraformer\_zh, indicating its strong capability in handling phonetic and lexical variations across dialects. Similar trends can be observed on challenging dialects such as Minnan and Hunan, where Dolphin-CN-Dialect consistently maintains robust performance despite significant pronunciation divergence from standard Mandarin.

Compared to larger-scale models, Dolphin-CN-Dialect remains highly competitive given its significantly smaller parameter size. While models such as FireRedASR-AED and FireRedASR-LLM achieve lower CER on certain dialects due to their larger capacity, Dolphin-CN-Dialect achieves comparable results on many subsets (e.g., Shanghai, Gansu, and Wenzhou), demonstrating an effective balance between model efficiency and dialect robustness.

The results on open-domain benchmarks further validate the generalization ability of Dolphin-CN-Dialect. As shown in Table~\ref{tab:kespeech_results}, Dolphin-CN-Dialect achieves substantial gains on both KeSpeech and Common Voice (tw), outperforming all sub-1B models by a clear margin. In particular, Dolphin-CN-Dialect achieves 5.04\% CER on KeSpeech and 5.62\% on Common Voice (tw), highlighting its robustness to diverse acoustic conditions and regional accent variations.

Overall, these results indicate that Dolphin-CN-Dialect effectively captures both standard Mandarin and dialectal characteristics through large-scale and diverse training data. This enables strong generalization across a wide spectrum of real-world speech scenarios, including dialectal variation, spontaneous speech, and noisy environments. Moreover, Dolphin-CN-Dialect significantly narrows the performance gap between sub-1B and billion-scale models on challenging dialect recognition tasks, making it well-suited for practical ASR deployment in multilingual and multi-dialect settings.

\subsection{Hotword Biasing Evaluation}

\begin{table*}[t]
\centering
\caption{Encoder-level contextual biasing results. Each entry reports WER (BWER $\mid$ UWER). RER denotes relative error reduction (\%) computed from w/o to w/ hotwords under the same decoding strategy.}
\label{tab:encoder_biasing}
\resizebox{0.95\linewidth}{!}{
\begin{tabular}{llcccc}
\toprule
Dataset & Model & Decoding & w/o hotwords & w/ hotwords & RER (\%) \\
\midrule

\multirow{4}{*}{AISHELL}
& Paraformer\_zh 
& - 
& 1.94 (18.76 $\mid$ 1.48) 
& 1.64 (6.42 $\mid$ 1.51) 
& \textbf{15.5} (\textbf{65.8} $\mid$ \textbf{-2.0}) \\

& Dolphin-CN-Dialect-0.4B 
& attention 
& 1.20 (10.77 $\mid$ 0.94) 
& \textbf{1.09} (4.85 $\mid$ \textbf{0.99}) 
& 9.20 (55.0 $\mid$ -5.3) \\

& Dolphin-CN-Dialect-0.4B 
& attention\_rescoring 
& 1.23 (11.31 $\mid$ 0.96) 
& 1.10 (\textbf{4.82} $\mid$ 1.00) 
& 10.6 (57.4 $\mid$ -4.2) \\

\midrule

\multirow{4}{*}{CommonVoice}
& Paraformer\_zh 
& - 
& 10.14 (22.96 $\mid$ 8.43) 
& 9.55 (18.15 $\mid$ 8.41) 
& 5.8 (20.9 $\mid$ \textbf{0.2}) \\

& Dolphin-CN-Dialect-0.4B 
& attention 
& 7.54 (16.63 $\mid$ 6.34) 
& \textbf{6.95} (10.18 $\mid$ \textbf{6.52}) 
& 7.8 (38.8 $\mid$ -2.8) \\

& Dolphin-CN-Dialect-0.4B 
& attention\_rescoring 
& 7.63 (16.17 $\mid$ 6.51) 
& 7.03 (\textbf{9.46} $\mid$ 6.71) 
& \textbf{7.9} (\textbf{41.5} $\mid$ -3.1) \\

\bottomrule
\end{tabular}
}
\end{table*}

To evaluate the effectiveness of the proposed hotword biasing methods, we construct two dedicated test sets with large-scale hotword annotations. These datasets are designed to cover both controlled conditions and long-tail lexical scenarios, enabling a comprehensive assessment of biasing performance.

\textbf{AISHELL Hotword Test Set.}
The AISHELL-based test set is constructed following the methodology described in prior work. Specifically, we adopt the hotword list released in \cite{shi2023seacoparaformernonautoregressiveasrflexible}, which contains 400 predefined hotwords. This dataset represents a relatively clean and controlled evaluation setting, where hotwords are moderately frequent and well-formed.

\textbf{CommonVoice Hotword Test Set.}
We further construct a large-scale hotword test set based on the Chinese subset of CommonVoice. Its main challenge lies in the \textbf{long-tail distribution of hotwords}. We automatically extract named entities and key phrases from the transcripts using a large language model, resulting in a total of 3,180 hotwords. Compared with AISHELL, these hotwords are significantly more diverse, less frequent, and often include translated or uncommon expressions, making the biasing task more challenging at the lexical level.

\textbf{Evaluation Protocol.}
We report three metrics to comprehensively evaluate hotword biasing performance:
(1) Word Error Rate (WER), measuring overall recognition accuracy;
(2) Biased Word Error Rate (BWER), focusing on recognition performance of hotwords;
and (3) Unbiased Word Error Rate (UWER), measuring the impact of biasing on non-hotword tokens.

\subsubsection{Encoder-Level Contextual Biasing}

We note that in CPPN-style\cite{huang2023contextualizedendtoendspeechrecognition} contextual biasing models, the number of sampled hotwords is not strictly fixed, but is implicitly constrained by the batch size and the sampling strategy (e.g., maximum sampling limits and per-utterance selection). In practice, when the batch size is small, the resulting hotword list remains relatively limited, and the mismatch between training and inference is less pronounced. However, as the batch size increases, more hotwords are aggregated across utterances, leading to significantly larger hotword lists after filtering. This results in a distribution shift, since the model is primarily trained on moderate-sized hotword sets but is exposed to much larger or much smaller sets during inference.

To address this issue, we adopt a dynamic batching strategy during training, where the number of hotwords per batch is explicitly varied over a wide range (from a few entries to several hundred). This allows the model to better generalize across different hotword list sizes, improving robustness under both dense and sparse biasing conditions.

To ensure fair comparison across different datasets, we fix the hotword filtering threshold to $-4$ and set the biasing weight to $0.5$ for experiments in this section. For the baseline system, we use Paraformer\_zh\cite{shi2023seacoparaformernonautoregressiveasrflexible} with its default decoding configuration ($nfilter=5$, $seaco\_weight=1.0$).

\begin{table*}[t]
\centering
\caption{Prompt-based hotword biasing results. Each entry reports WER (BWER $\mid$ UWER). RER denotes relative error reduction (\%).}
\label{tab:prompt_biasing}
\resizebox{0.95\linewidth}{!}{
\begin{tabular}{llccc}
\toprule
Dataset & Model & w/o hotwords & w/ hotwords & RER (\%) \\
\midrule

\multirow{2}{*}{AISHELL}
& Fun-ASR-Nano-2512 
& {1.65 (14.01 $\mid$ 1.32) }
& {1.45 (6.39 $\mid$ 1.31)} 
& {\textbf{12.1} (\textbf{54.4} $\mid$ \textbf{0.8})} \\

& Dolphin-CN-Dialect-0.4B-ft-prompt 
& {1.26 (11.90 $\mid$ 0.97)}
& {\textbf{1.11} (\textbf{5.99} $\mid$ \textbf{0.97})}
& {11.9 (50.0 $\mid$ 0.0)} \\

\midrule

\multirow{2}{*}{CommonVoice}
& Fun-ASR-Nano-2512  
& {6.76 (14.37 $\mid$ 5.76)} 
& {6.30 (7.15 $\mid$ 6.19) }
& {6.8 (50.2 $\mid$ -7.5)} \\

& Dolphin-CN-Dialect-0.4B-ft-prompt 
& {7.11 (15.22 $\mid$ 6.04)} 
& {\textbf{6.08} (\textbf{6.79} $\mid$ \textbf{5.99}) }
& {\textbf{14.5} (\textbf{55.4} $\mid$ \textbf{0.8})} \\

\bottomrule
\end{tabular}
}
\end{table*}

\begin{table*}[t]
\centering
\caption{Oracle prompt-based hotword biasing results. Each entry reports WER (BWER $\mid$ UWER). RER denotes relative error reduction (\%) computed for each metric.}
\label{tab:oracle_prompt_biasing}
\resizebox{0.95\linewidth}{!}{
\begin{tabular}{
ll
S[table-format=2.2]
S[table-format=2.2]
S[table-format=2.1]
}
\toprule
Dataset & Model & {w/o hotwords} & {w/ hotwords} & {RER (\%)} \\
\midrule

\multirow{2}{*}{AISHELL}
& Fun-ASR-Nano-2512
& {4.19 (14.05 $\mid$ 1.53)} 
& {1.83 (2.81 $\mid$ 1.57)} 
& {56.3 (80.0 $\mid$ -2.6)} \\

& Dolphin-CN-Dialect-0.4B-ft-prompt 
& {3.56 (11.93 $\mid$ 1.30)} 
& {\textbf{1.26} (\textbf{0.47} $\mid$ \textbf{1.09})} 
& {\textbf{64.6} (\textbf{96.1} $\mid$ \textbf{16.2})} \\

\midrule

\multirow{2}{*}{CommonVoice}
& Fun-ASR-Nano-2512
& {8.96 (14.33 $\mid$ 5.19)} 
& {4.46 (3.37 $\mid$ \textbf{5.22})} 
& {50.2 (76.5 $\mid$ -0.6)} \\

& Dolphin-CN-Dialect-0.4B-ft-prompt 
& {9.60 (15.15 $\mid$ 5.70)} 
& {\textbf{4.14} (\textbf{2.33} $\mid$ 5.41)} 
& {\textbf{56.9} (\textbf{84.6} $\mid$ \textbf{5.1})} \\

\bottomrule
\end{tabular}
}
\end{table*}

Overall, encoder-level biasing significantly improves hotword recognition across all models, as reflected by substantial reductions in BWER. Notably, Dolphin consistently achieves lower BWER than Paraformer\_zh under both datasets and decoding strategies. For example, on AISHELL, Dolphin reduces BWER to 4.85 (attention) and 4.82 (rescoring), compared to 6.42 for Paraformer\_zh. On CommonVoice, the gap is more pronounced, where Dolphin achieves 9.46 BWER with rescoring, significantly outperforming Paraformer\_zh (18.15). These results demonstrate that Dolphin is more effective at incorporating contextual information for accurate hotword recognition.

In terms of relative improvements, Dolphin also achieves competitive or higher BWER reduction compared to Paraformer\_zh. For instance, on CommonVoice, Dolphin achieves up to 41.5\% relative reduction in BWER, compared to 20.9\% for Paraformer\_zh. This suggests that Dolphin not only starts from a stronger baseline but also benefits more from contextual biasing, especially in challenging long-tail scenarios.

In addition to hotword recognition, Dolphin maintains better overall accuracy. On AISHELL, Dolphin consistently achieves lower WER than Paraformer\_zh (e.g., 1.09 vs. 1.64), indicating that the improvements in BWER translate into gains in overall recognition. Similar trends are observed on CommonVoice.

We further analyze the impact on non-hotword tokens. Dolphin exhibits a slight increase in UWER after biasing (e.g., 0.94 → 0.99 on AISHELL), indicating a mild trade-off between hotword recall and general recognition. In contrast, Paraformer\_zh shows more stable UWER, suggesting a more conservative biasing behavior. Nevertheless, the degradation for Dolphin remains limited compared to the substantial gains in BWER.

Finally, comparing decoding strategies, attention rescoring consistently yields slightly better BWER than attention decoding for Dolphin (e.g., 4.85 → 4.82 on AISHELL and 10.18 → 9.46 on CommonVoice), further enhancing hotword recall. The corresponding changes in WER and UWER remain minor, indicating that encoder-level biasing in Dolphin is robust to decoding variations.

Overall, these results show that Dolphin provides more effective contextual biasing than Paraformer\_zh, achieving significantly better hotword recognition while maintaining competitive overall performance.

\subsubsection{Prompt-Based Hotword Biasing}

We further evaluate the prompt-based hotword biasing method described in Section 2.3.2. In this setting, hotwords are directly incorporated into the decoder input as contextual prompts.

We use Dolphin-CN-Dialect-0.4B-ft-prompt, which is fine-tuned with the proposed prompt-based training strategy. As a baseline, we adopt Fun-ASR-Nano-2512\cite{an2025fun}, which also introduces hotwords as prompts during decoding.

A key challenge in prompt-based biasing is that directly injecting a large number of hotwords can severely degrade decoding quality. Since Fun-ASR-Nano-2512 does not include a built-in hotword filtering mechanism, we apply a unified \textit{Contextual Phrase Filtering}\cite{huang2023contextualizedendtoendspeechrecognition} strategy to both models. Specifically, we first filter the candidate hotword list using a confidence threshold of $-2$, and then feed the filtered hotwords as prompts into both models. This ensures a fair and stable comparison. Table~\ref{tab:prompt_biasing} presents the results on both AISHELL and CommonVoice test sets. Only \textit{attention beam seach} results are presented.

Overall, prompt-based biasing significantly improves hotword recognition for both models, as indicated by substantial reductions in BWER. However, Dolphin-CN-Dialect-0.4B-ft-prompt consistently achieves better performance than Fun-ASR-Nano-2512 in terms of absolute accuracy and robustness. On AISHELL, Dolphin-CN-Dialect-0.4B-ft-prompt reduces BWER to 5.99, outperforming Fun-ASR-Nano-2512 (6.39), while also achieving lower WER (1.11 vs. 1.45). On CommonVoice, the advantage becomes more pronounced, where Dolphin-CN-Dialect-0.4B-ft-prompt achieves 6.79 BWER compared to 7.15 for Fun-ASR-Nano, together with a lower WER (6.08 vs. 6.30). These results indicate that Dolphin-CN-Dialect-0.4B-ft-prompt can more effectively leverage prompt information for accurate hotword recognition.

In terms of relative improvements, Dolphin-CN-Dialect-0.4B-ft-prompt also demonstrates stronger gains, particularly on more challenging datasets. In CommonVoice, Dolphin-CN-Dialect-0.4B-ft-prompt achieves a relative reduction of 55.4\% in BWER, surpassing Fun-ASR-Nano-2512 (50.2\%), and a significantly larger reduction in WER (14.5\% vs. 6.8\%). This suggests that Dolphin-CN-Dialect-0.4B-ft-prompt benefits more from prompt-based biasing, especially in long-tail scenarios where hotword coverage is more difficult.

We further analyze the impact on non-hotword tokens. Dolphin-CN-Dialect-0.4B-ft-prompt maintains stable UWER across both datasets (e.g., 0.97\% → 0.97\% on AISHELL and 6.04\% → 5.99\% on CommonVoice), indicating that prompt-based biasing introduces minimal side effects. In contrast, Fun-ASR-Nano-2512 exhibits noticeable degradation on CommonVoice (5.76\% → 6.19\%), suggesting that it is more sensitive to prompt injection and may suffer from reduced general recognition accuracy.

Overall, these results demonstrate that Dolphin-CN-Dialect-0.4B-ft-prompt provides more effective and stable prompt-based biasing than Fun-ASR-Nano-2512, achieving better hotword recognition while preserving general ASR performance.

\subsubsection{Oracle Analysis of Prompt-Based Hotword Biasing}

In practical systems, prompt-based hotword biasing depends on an upstream hotword filtering module, which may introduce errors. To isolate the intrinsic effectiveness of prompt-based biasing, we conducted an oracle experiment in which only ground-truth hotwords are provided.

We construct evaluation subsets from AISHELL and CommonVoice by selecting utterances that contain predefined hotwords. The AISHELL subset consists of 808 utterances with 400 hotwords, while the CommonVoice subset contains 2,602 utterances with 3,180 hotwords. During decoding, the exact ground-truth hotwords for each utterance are used as prompts, removing any noise or recall issues from hotword selection. The results are summarized in Table~\ref{tab:oracle_prompt_biasing}. Only \textit{attention beam seach} results are presented.

On AISHELL, prompt-based biasing leads to a substantial improvement for Dolphin-CN-Dialect-0.4B-ft-prompt  under attention decoding, reducing WER from 3.56 to 1.26(relative improvement \textbf{64.6\%}) and BWER from 11.93 to 0.47(relative improvement \textbf{96\%}) . This indicates that, when correct hotwords are provided, the model is able to almost perfectly recognize target keywords. Similar trends are observed on CommonVoice, where WER is reduced from 9.60 to 4.14(relative improvement \textbf{56.8\%}) and BWER from 15.15 to 2.33(relative improvement \textbf{84.6\%}), demonstrating strong effectiveness even in long-tail and noisy scenarios.

Compared with Fun-ASR-Nano-2512, Dolphin achieves consistently lower BWER under the oracle setting, indicating a stronger ability to leverage prompt information for hotword recognition, particularly in long-tail scenarios.

For Dolphin-CN-Dialect-0.4B-ft-prompt , UWER is slightly reduced after introducing oracle hotwords (e.g., 1.30 → 1.09 on AISHELL and 5.70 → 5.41 on CommonVoice), indicating that prompt-based biasing does not harm, and may even slightly improve, general recognition.

In contrast, Fun-ASR-Nano-2512 shows a slight increase in UWER (e.g., 1.53 → 1.57 on AISHELL and 5.19 → 5.22 on CommonVoice), suggesting that hotword biasing introduces mild degradation on non-hotword tokens.

\section{Discussion}

\subsection{Data-Centric ASR Design}

Our results highlight the importance of data-centric approaches in modern ASR systems. While model architecture remains important, many of the performance gains in Dolphin-CN-Dialect come from improvements in data processing, sampling strategies, and tokenizer design. In particular, the temperature-based sampling method demonstrates that carefully reshaping the training distribution can significantly improve performance on low-resource dialects without sacrificing overall accuracy.

\subsection{Dialect Modeling and Long-Tail Distribution}

Dolphin-CN-Dialect provides evidence that addressing the long-tail distribution problem is essential for building robust multilingual ASR systems. Standard training pipelines tend to favor high-resource languages, leading to poor generalization in dialect-rich scenarios. By explicitly modeling dialects through both sampling strategies and token design, Dolphin-CN-Dialect achieves substantial improvements in dialect recognition. This suggests that future ASR systems should incorporate more explicit mechanisms for handling linguistic diversity.

\subsection{Tokenizer Design for Multilingual ASR}

The redesigned tokenizer in Dolphin-CN-Dialect reflects the importance of aligning tokenization strategies with linguistic properties. The combination of character-level modeling for Chinese and subword modeling for alphabetic languages enables better alignment and representation efficiency. Furthermore, the introduction of extensible dialect tokens provides a scalable path toward finer-grained language modeling. This hybrid approach may serve as a useful reference for future multilingual ASR systems.

\subsection{Bridging Research and Deployment}

Beyond model performance, Dolphin-CN-Dialect emphasizes practical considerations such as training stability, streaming capability, and system efficiency. The engineering challenges we encountered demonstrate that real-world ASR systems require careful integration of algorithms and infrastructure. Improvements such as data pipeline optimization, and memory-aware training are critical for scaling models in production environments.

\subsection{Limitations and Future Directions}

Despite these improvements, several limitations remain. First, while the temperature-based sampling strategy improves dialect performance, it still relies on manual tuning of the hyperparameter $\alpha$. Second, Dolphin-CN-Dialect currently focuses primarily on Chinese and dialect-rich scenarios, and its performance on broader multilingual benchmarks requires further evaluation. Finally, although streaming and non-streaming modes are supported, achieving optimal trade-offs between latency and accuracy remains an open problem.

Future work will explore more adaptive data balancing strategies, improved streaming architectures, and further expansion to low-resource languages and dialects.

\section{Conclusion}

In this work, we presented Dolphin-CN-Dialect, a multi-dialect ASR model designed for real-world speech recognition scenarios. Building upon Dolphin, Dolphin-CN-Dialect introduces a series of improvements in tokenizer design, data sampling strategy, training stability, and system efficiency.

We propose a temperature-based sampling method to address the long-tail distribution of dialect data, significantly improving recognition performance for low-resource dialects. Combined with a redesigned tokenizer and enhanced data processing pipeline, Dolphin-CN-Dialect achieves a 38\% improvement in dialect accuracy and a 16\% relative reduction in overall CER compared to the previous version.

In addition, Dolphin-CN-Dialect supports both streaming and non-streaming inference, enabling flexible deployment across a wide range of applications. The model also demonstrates competitive performance with recent open-source ASR systems, while maintaining a more efficient model size.

Overall, Dolphin-CN-Dialect highlights the importance of data-centric design and system-level optimization in modern ASR. We hope this work provides useful insights for building robust, scalable, and practical multilingual speech recognition systems.

\section{Generative AI Use Disclosure}

This work utilized generative AI tools to assist in drafting and refining parts of the manuscript, including language polishing and structural organization. All technical content, experimental design, model development, and results are solely the responsibility of the authors.

The authors have carefully reviewed and verified all AI-assisted content to ensure its accuracy, correctness, and consistency with the research contributions presented in this paper.

\bibliographystyle{IEEEtran}
\bibliography{tech_reportv2}

\end{document}